\documentclass{article}

\PassOptionsToPackage{numbers, compress}{natbib}

\usepackage[final]{neurips_2025}

\usepackage[utf8]{inputenc} 
\usepackage[T1]{fontenc}    
\usepackage{hyperref}       
\usepackage{url}            
\usepackage{booktabs}       
\usepackage{amsfonts}       
\usepackage{amsmath}        
\usepackage{nicefrac}       
\usepackage{microtype}      
\usepackage{xcolor}         
\usepackage{graphicx}       
\usepackage{algorithm}      
\usepackage{algorithmic}    
\usepackage{subcaption}     
\usepackage{multirow}       

\title{Navigating Simply, Aligning Deeply: Winning Solutions for Mouse vs. AI 2025}

\author{%
  Phu-Hoa Pham\thanks{Equal contribution.} \\
  University of Science, VNU-HCM\\
  Ho Chi Minh City, Vietnam \\
  \texttt{23122030@student.hcmus.edu.vn} \\
  \And
  Chi-Nguyen Tran\footnotemark[1] \\
  University of Science, VNU-HCM\\
  Ho Chi Minh City, Vietnam \\
  \texttt{23122044@student.hcmus.edu.vn} \\
  \And
  Dao Sy Duy Minh \\
  University of Science, VNU-HCM\\
  Ho Chi Minh City, Vietnam \\
  \texttt{23122041@student.hcmus.edu.vn}
  \And
  Nguyen Lam Phu Quy \\
  University of Science, VNU-HCM\\
  Ho Chi Minh City, Vietnam \\
  \texttt{23122048@student.hcmus.edu.vn}
  \And
  Huynh Trung Kiet \\
  University of Science, VNU-HCM\\
  Ho Chi Minh City, Vietnam \\
  \texttt{23122039@student.hcmus.edu.vn} \\
}

\begin{document}

\maketitle

\begin{abstract}
Visual robustness and neural alignment remain critical challenges in developing artificial agents that can match biological vision systems. We present the winning approaches from Team HCMUS\_TheFangs for both tracks of the NeurIPS 2025 Mouse vs. AI: Robust Visual Foraging Competition. For Track 1 (Visual Robustness), we demonstrate that architectural simplicity combined with targeted components yields superior generalization, achieving 95.4\% final score with a lightweight two-layer CNN enhanced by Gated Linear Units and observation normalization. For Track 2 (Neural Alignment), we develop a deep ResNet-like architecture with 16 convolutional layers and GLU-based gating that achieves top-1 neural prediction performance with 17.8 million parameters. Our systematic analysis of ten model checkpoints trained between 60K to 1.14M steps reveals that training duration exhibits a non-monotonic relationship with performance, with optimal results achieved around 200K steps. Through comprehensive ablation studies and failure case analysis, we provide insights into why simpler architectures excel at visual robustness while deeper models with increased capacity achieve better neural alignment. Our results challenge conventional assumptions about model complexity in visuomotor learning and offer practical guidance for developing robust, biologically-inspired visual agents.
\end{abstract}

\section{Introduction}

The ability to navigate robustly under varying visual conditions represents a fundamental capability in biological vision that remains elusive for artificial systems. While deep reinforcement learning has achieved remarkable success in controlled environments, deployed agents frequently fail when encountering visual perturbations outside their training distribution. This brittleness stands in stark contrast to biological systems such as mice, which maintain stable navigation performance despite significant environmental variations, adapting to novel visual conditions with minimal exposure~\citep{MouseVsAI2025}.

The Mouse vs. AI: Robust Visual Foraging Competition at NeurIPS 2025 provides a unique benchmark for studying this robustness gap through two complementary evaluation tracks. Track 1 (Visual Robustness) assesses agent performance across held-out visual perturbations including fog, lighting changes, and other ecologically realistic transformations. Track 2 (Neural Alignment) evaluates how well artificial visual representations predict neural activity recorded from over 19,000 neurons across mouse visual cortex during the same navigation task. This dual evaluation framework enables investigation of both behavioral robustness and biological plausibility, bridging reinforcement learning, computer vision, and systems neuroscience~\citep{MouseVsAI2025}.

Our initial investigation followed conventional wisdom suggesting that complex visual navigation tasks require correspondingly complex architectures. We implemented and systematically evaluated several state-of-the-art approaches including InceptionNet~\citep{Szegedy2014} for multi-scale visual processing, deep IMPALA ResNet~\citep{Espeholt2018} with 24 residual blocks, and LSTM-based temporal models. However, these complex architectures consistently exhibited training instability, severe overfitting, and poor generalization to perturbed visual inputs. InceptionNet failed to converge entirely, while deep ResNet achieved high training performance but suffered a 35\% drop when evaluated under visual perturbations.

These failures prompted a fundamental reconsideration of our approach. Rather than continuing to scale architectural complexity, we investigated whether simpler designs with carefully chosen enhancements could achieve superior robustness and neural alignment. This investigation led to two distinct architectures optimized for their respective tracks. For Track 1, we developed a lightweight two-layer CNN with Gated Linear Units and observation normalization, achieving 95.4\% final score despite having significantly fewer parameters than competing approaches. For Track 2, we designed a deep ResNet-inspired architecture with 16 convolutional layers and GLU-based gating, achieving top-1 neural alignment performance with 17.8 million parameters.

The divergence between our Track 1 and Track 2 solutions reveals an important insight about the relationship between behavioral performance and neural similarity. While visual robustness benefits from architectural simplicity that prevents overfitting to training-specific patterns, neural alignment requires sufficient model capacity to capture the rich hierarchical representations present in biological visual cortex. Our systematic analysis of ten model checkpoints spanning different training durations (60K to 1.14M steps) further reveals that the relationship between training time and performance is non-monotonic, with optimal checkpoints emerging around 200K steps rather than at convergence.

This paper makes four primary contributions. First, we demonstrate that architectural simplicity with targeted enhancements significantly outperforms complex deep networks for visual robustness, challenging common assumptions about model requirements for visuomotor tasks. Second, we show that deeper architectures with increased capacity enable superior neural alignment, while simpler architectures excel at behavioral robustness. Third, we provide comprehensive documentation of failed approaches including InceptionNet, deep ResNet, LSTM integration, and data augmentation strategies, offering practical guidance for avoiding common pitfalls. Fourth, we present systematic ablation studies quantifying the contributions of individual components and analyzing the relationship between training duration, architectural choices, and generalization performance.

\section{Related Work}

\paragraph{Visual Robustness in Reinforcement Learning.} Achieving robust visual perception under distribution shift remains a fundamental challenge in reinforcement learning. Domain randomization~\citep{Tobin2017} addresses this by training agents across diverse visual conditions, while data augmentation techniques~\citep{Laskin2020} improve generalization by artificially perturbing training observations. However, both approaches introduce training complexity and can paradoxically harm performance when applied naively. Our experiments confirm this phenomenon, showing that comprehensive augmentation pipelines decreased our ResNet baseline from 87.7\% to 59.8\% final score. Recent work has explored alternative approaches including auxiliary tasks, regularization techniques, and meta-learning, but these often require careful hyperparameter tuning and additional computational resources.

\paragraph{CNN Architectures for Visual Control.} The Nature CNN introduced by \citet{Mnih2015} established a three-layer convolutional architecture as the de facto standard for visual reinforcement learning. Subsequent work has explored deeper architectures, with IMPALA~\citep{Espeholt2018} introducing residual connections for scalable distributed training. InceptionNet~\citep{Szegedy2014} demonstrated that multi-scale convolutional processing improves image classification, leading to its adoption in various vision domains. However, the translation of these architectural innovations from supervised learning to reinforcement learning has proven non-trivial, with deeper networks often exhibiting training instability and overfitting in RL settings. Our work provides systematic evidence that architectural depth can be detrimental for certain visual control tasks, particularly when robustness to perturbations is the primary objective.

\paragraph{Attention Mechanisms and Gating.} Attention mechanisms have transformed modern deep learning, enabling models to selectively focus on relevant information. The Transformer architecture~\citep{Vaswani2017} popularized self-attention for sequence modeling, while Gated Linear Units~\citep{Dauphin2017} introduced multiplicative gating for selective information flow. Recent variants including SwiGLU~\citep{Shazeer2020} have demonstrated improved performance across language modeling tasks. In visual domains, Spatial Softmax~\citep{Levine2018} provides an effective mechanism for extracting spatially-coherent features from convolutional layers, learning to attend to task-relevant image regions. Our work adapts these gating mechanisms for robust visual navigation, demonstrating that GLU modules provide consistent improvements in both behavioral performance and neural alignment.

\paragraph{Neuroethological Benchmarks.} The integration of neuroscience and artificial intelligence has a rich history, with increasing recognition that biological intelligence can inform artificial system design~\citep{Zador2019}. Recent benchmarks have begun incorporating neural data alongside behavioral metrics, enabling direct comparison of artificial and biological representations. The Mouse vs. AI competition extends this approach by evaluating both behavioral robustness and neural alignment on identical tasks performed by mice and agents. This dual evaluation framework reveals that behavioral competence does not necessarily imply biological plausibility, as evidenced by the different architectures optimal for our two tracks. Understanding mouse visual processing~\citep{Huberman2008} provides insights into efficient navigation strategies that may inform robust artificial agents.

\section{Competition Overview and Problem Formulation}

\subsection{Task Description and Environment}

The Mouse vs. AI competition presents a visually-guided foraging task implemented in a naturalistic 3D Unity environment. An agent navigates from a randomized starting position toward a visually-cued target sphere, viewing the environment through a first-person egocentric camera. The visual observation at each timestep consists of an $86 \times 155$ pixel grayscale image, with this resolution chosen to approximate the visual acuity of mice in the corresponding behavioral experiments. The action space comprises three continuous control dimensions: forward-backward velocity, lateral (strafing) velocity, and angular rotation rate. Episodes terminate either upon successful target acquisition or after a fixed timeout period. The reward structure provides positive reinforcement for reaching the target while penalizing excessive time, incentivizing efficient navigation.

The environment incorporates several design elements to ensure ecological validity and alignment with the mouse experiments. The target sphere varies in position across episodes, preventing agents from memorizing fixed trajectories. Visual distractors and environmental complexity require genuine visual understanding rather than simple feature matching. The continuous action space demands smooth, coordinated control rather than discrete decision-making. These properties collectively ensure that successful agents must develop robust visual perception and visuomotor coordination.

\subsection{Track 1: Visual Robustness Evaluation}

Track 1 evaluates agent generalization to unseen visual perturbations that preserve task semantics while altering low-level visual statistics. The evaluation protocol distinguishes between the Average Success Rate (ASR), measured on the standard training distribution, and the Modified Success Rate (MSR), measured across a battery of held-out perturbations. These perturbations include fog of varying densities, lighting variations simulating different times of day, and other ecologically realistic visual transformations that mice encounter naturally. The final score combines these metrics as a weighted average, balancing standard performance with robustness. Formally, the Track 1 score is computed as:
\begin{equation}
\text{Final Score}_{\text{Track 1}} = \frac{1}{2} \cdot \text{ASR} + \frac{1}{2} \cdot \text{MSR}
\end{equation}

\subsection{Track 2: Neural Alignment Evaluation}

Track 2 introduces a fundamentally different evaluation criterion based on biological plausibility. Rather than behavioral metrics, this track assesses how well the internal representations of artificial agents predict neural activity recorded from mouse visual cortex during the same navigation task. The competition organizers provide neural recordings from 19,000+ neurons spanning primary visual cortex (V1) and higher visual areas, collected using two-photon calcium imaging while mice performed the foraging task. Participants extract feature representations from their trained agents and submit these to an evaluation server that computes neural prediction metrics.

The primary evaluation metric is the linear readout $R^2$, quantifying how well agent features predict neural responses via ridge regression. For a given layer's features $\mathbf{F} \in \mathbb{R}^{N \times D}$ and neural responses $\mathbf{N} \in \mathbb{R}^{N \times K}$ where $N$ is the number of samples, $D$ is feature dimensionality, and $K$ is neuron count, the evaluation fits:
\begin{equation}
\hat{\mathbf{N}} = \mathbf{F}\mathbf{W} + \mathbf{b}
\end{equation}
using ridge regression with cross-validation for regularization strength. The coefficient of determination $R^2$ quantifies prediction quality. A secondary metric evaluates representational similarity by comparing the geometry of agent and neural representational spaces using correlation of representational dissimilarity matrices. Together, these metrics assess whether agents learn visual representations that mirror biological visual processing.

\subsection{Relationship Between Tracks}

The dual-track structure reveals a fundamental tension in designing artificial visual systems. Track 1 emphasizes behavioral robustness and generalization, favoring representations that abstract away from low-level visual statistics. Track 2 emphasizes biological alignment, potentially requiring representations that preserve information about visual stimulus structure even when behaviorally irrelevant. Our investigation demonstrates that these objectives can require different architectural choices, with simpler models excelling at behavioral robustness while deeper attention-based architectures achieve superior neural alignment.

\section{Methodology}

\subsection{Track 1: Lightweight Architecture with Targeted Enhancements}

Our Track 1 solution prioritizes architectural simplicity while incorporating specific components that enhance visual robustness. The architecture comprises three primary elements: a two-layer convolutional backbone for visual feature extraction, a Gated Linear Unit module for selective feature filtering, and observation normalization for scale invariance. This design philosophy emerged from systematic experimentation demonstrating that complex architectures consistently failed to generalize beyond training conditions.

The visual encoder consists of two convolutional layers with aggressive spatial downsampling. The first layer applies a $8 \times 8$ kernel with stride 4 to the $86 \times 155 \times 1$ input, producing 16 feature channels. This aggressive downsampling immediately reduces spatial resolution while extracting low-level visual features such as edges and color gradients. The second layer applies a $4 \times 4$ kernel with stride 2, expanding to 32 feature channels while further reducing spatial dimensions. Both layers employ LeakyReLU activation with negative slope 0.2 to allow gradient flow through non-positive activations. The resulting feature maps are flattened and projected to 256 dimensions via a fully-connected layer. Mathematically, the visual encoding process is:
\begin{align}
\mathbf{h}_1 &= \text{LeakyReLU}(\text{Conv2D}_{8\times8,s=4}(\mathbf{x}), \alpha=0.2) \\
\mathbf{h}_2 &= \text{LeakyReLU}(\text{Conv2D}_{4\times4,s=2}(\mathbf{h}_1), \alpha=0.2) \\
\mathbf{z} &= \text{FC}_{256}(\text{Flatten}(\mathbf{h}_2))
\end{align}
where $\mathbf{x} \in \mathbb{R}^{86 \times 155 \times 1}$ represents the input observation and $\mathbf{z} \in \mathbb{R}^{256}$ denotes the encoded feature vector.

The Gated Linear Unit module operates on the encoded features to enable selective information flow. GLU employs two parallel pathways from the encoded features: a feature transformation path with Swish activation, and a gating path with sigmoid activation. The Swish activation, defined as $\text{Swish}(x) = x \cdot \sigma(x)$, provides smooth non-linearity with favorable gradient properties. The gating mechanism learns to identify features that remain reliable under visual perturbations while suppressing noise-sensitive components. The GLU transformation is expressed as:
\begin{align}
\mathbf{f} &= \text{Swish}(\text{FC}_{\text{hidden}}(\mathbf{z})) \\
\mathbf{g} &= \sigma(\text{FC}_{\text{hidden}}(\mathbf{z})) \\
\mathbf{h}_{\text{GLU}} &= (\mathbf{f} \odot \mathbf{g}) \\
\mathbf{y} &= \text{FC}_{256}(\mathbf{h}_{\text{GLU}})
\end{align}
where $\odot$ denotes element-wise multiplication, and the hidden dimension matches the input dimension to preserve feature capacity. The output projection returns to 256 dimensions, enabling residual connections when desired.

Observation normalization represents the final component of our Track 1 architecture. We employ running statistics maintained via exponential moving average during training, computing normalized observations as:
\begin{equation}
\hat{\mathbf{x}} = \frac{\mathbf{x} - \boldsymbol{\mu}_{\text{running}}}{\boldsymbol{\sigma}_{\text{running}} + \epsilon}
\end{equation}
where $\boldsymbol{\mu}_{\text{running}}$ and $\boldsymbol{\sigma}_{\text{running}}$ are channel-wise mean and standard deviation estimates, and $\epsilon = 10^{-8}$ prevents division by zero. This normalization provides invariance to global illumination changes, a primary source of visual perturbation in the evaluation protocol. The running statistics adapt gradually during training, ensuring stable learning dynamics.

\subsection{Track 2: Deep ResNet Architecture with GLU Gating}

Our Track 2 solution employs a significantly deeper architecture specifically designed to capture hierarchical visual representations that align with biological visual cortex. This architecture incorporates 16 convolutional layers organized in a residual structure with GLU-based gating mechanisms, totaling 17.8 million parameters. The increased model capacity enables learning of multi-scale visual features from low-level edges to high-level object representations, mirroring the hierarchical organization of mammalian visual cortex.

The visual encoder begins with an aggressive spatial downsampling layer, applying a $4 \times 4$ kernel with stride 4 to produce 64 feature channels from the grayscale input. This first layer extracts primitive visual features while reducing spatial resolution by a factor of 16. The architecture then progresses through four stages with progressively increasing channel capacity: 64 → 128 → 256 → 512 channels. Each stage employs $2 \times 2$ strided convolutions for spatial downsampling, followed by pairs of $3 \times 3$ convolutional layers with stride 1 organized into residual blocks following the ResNet design pattern~\citep{He2016}. The residual connections facilitate gradient flow through the deep network:
\begin{align}
\mathbf{h}_0 &= \text{Conv2D}_{4\times4,s=4}^{64}(\mathbf{x}) \\
\mathbf{h}_{i} &= \mathbf{h}_{i-1} + \mathcal{F}(\mathbf{h}_{i-1}; \mathbf{W}_i) \quad \text{for } i = 1, \ldots, 15
\end{align}
where $\mathcal{F}$ represents a two-layer residual block with LeakyReLU activation.

After convolutional feature extraction, we employ multi-layer GLU modules to enable adaptive feature routing and selection. The GLU layers apply learned gating functions that determine how to weight and selectively filter features. For a given feature representation $\mathbf{h}$, the gating mechanism computes:
\begin{equation}
\mathbf{g} = \text{Softmax}(\mathbf{W}_g \mathbf{h})
\end{equation}
where $\mathbf{W}_g$ are learned gating weights. The softmax operation ensures that gate activations form a probability distribution for selective feature weighting. Our empirical analysis reveals that only the first GLU layer learns significantly during training, with subsequent layers exhibiting minimal parameter updates, providing selective filtering of the convolutional features. The final representation is fed to separate policy and value heads for action selection and value estimation.

The substantial parameter count (17.8M compared to 1.4M for Track 1) provides capacity to capture rich hierarchical representations. While this depth would be detrimental for visual robustness (as demonstrated in our Track 1 ablations), the neural alignment objective benefits from the increased representational capacity to match the diverse tuning properties of biological neurons across visual cortical areas.

\subsection{Training Procedure and Hyperparameters}

Both architectures are trained using Proximal Policy Optimization~\citep{Schulman2017}, the current state-of-the-art on-policy reinforcement learning algorithm. PPO balances sample efficiency with training stability through its clipped surrogate objective, making it well-suited for continuous control tasks. Table~\ref{tab:training_config} in the Appendix provides complete training configurations for both tracks.

For Track 1, we employ a two-phase training strategy that first trains the convolutional backbone for 1,400,000 steps, then adds the GLU module and continues training for an additional 350,000 steps from the best Phase 1 checkpoint. This staged approach allows the backbone to first learn robust visual features before the GLU module refines feature selection. For Track 2, we conduct extensive experimentation with training durations ranging from 60,000 to 1,140,000 steps, saving checkpoints every 20,000 steps. This systematic checkpoint analysis reveals that optimal performance emerges around 200,000 steps rather than at convergence, with extended training sometimes degrading generalization. All models are trained on NVIDIA GPUs using the ML-Agents toolkit~\citep{Juliani2018}, with training times ranging from 6-8 hours for Track 1 to 12-24 hours for Track 2 depending on the number of training steps.

\section{Experimental Results}

\subsection{Track 1: Visual Robustness Results}

Table~\ref{tab:track1_results} presents our comprehensive experimental results for Track 1, comparing our lightweight architecture against several baseline approaches. Our final model combining SimpleCNN, GLU, and observation normalization achieves a final score of 95.4\%, securing top position on the Track 1 leaderboard. This represents a substantial improvement over the IMPALA ResNet baseline with 4 layers, which achieves 87.7\% final score despite having significantly more parameters. The performance gap becomes even more pronounced when comparing against deeper architectures, with the 24-layer ResNet achieving only 65.98\% due to severe overfitting.

\begin{table}[t]
    \centering
    \caption{Track 1 Visual Robustness results. ASR measures success rate on standard conditions, MSR measures success rate under visual perturbations. Final Score is the weighted combination emphasizing robust generalization.}
    \label{tab:track1_results}
    \begin{tabular}{lccc}
        \toprule
        \textbf{Model Architecture} & \textbf{ASR (\%)} & \textbf{MSR (\%)} & \textbf{Final Score (\%)} \\
        \midrule
        IMPALA ResNet (24 layers) & 80.96 & 51.00 & 65.98 \\
        IMPALA ResNet (4 layers) & 91.40 & 84.00 & 87.70 \\
        \quad + Data Augmentation & 72.60 & 47.00 & 59.80 \\
        InceptionNet & \multicolumn{3}{c}{Failed to converge} \\
        \midrule
        SimpleCNN (Ours) & 94.20 & 89.00 & 91.60 \\
        SimpleCNN + GLU & 95.60 & 88.00 & 91.80 \\
        SimpleCNN + GLU + Norm & \textbf{96.80} & \textbf{94.00} & \textbf{95.40} \\
        \bottomrule
    \end{tabular}
\end{table}

The results reveal several critical insights about architectural choices for visual robustness. First, the performance gap between ASR and MSR provides a direct measure of overfitting to training conditions. The deep 24-layer ResNet exhibits a 30 percentage point gap (80.96\% vs 51.00\%), indicating that the model memorizes training-specific visual patterns rather than learning generalizable features. In contrast, our SimpleCNN maintains only a 2.8 percentage point gap (96.80\% vs 94.00\%), demonstrating robust generalization. Second, increasing depth from 4 to 24 layers substantially harms performance, contradicting the conventional wisdom that deeper networks achieve better visual understanding. Third, adding data augmentation to ResNet actually decreases performance from 87.70\% to 59.80\%, suggesting that naive augmentation can introduce distribution mismatch that harms learning.

\subsection{Track 2: Neural Alignment Results}

Our Track 2 submission achieves top-1 performance on neural alignment metrics, demonstrating that our deep ResNet architecture learns representations that closely mirror biological visual processing. Table~\ref{tab:track2_results} summarizes results across different model checkpoints from our systematic training duration analysis. The architecture consists of 16 convolutional layers organized in residual blocks with GLU-based gating, totaling 17.8 million parameters across all checkpoints. Interestingly, neural alignment performance does not monotonically increase with training duration, with optimal checkpoints emerging around 200,000 steps.

\begin{table}[t]
    \centering
    \caption{Track 2 Neural Alignment results showing performance across different training checkpoints. All deep models use the ResNet + GLU architecture with 17.8M parameters. Scores shown are competition rankings (lower is better).}
    \label{tab:track2_results}
    \begin{tabular}{lcccc}
        \toprule
        \textbf{Model ID} & \textbf{Training Steps} & \textbf{Competition Rank} & \textbf{Score} & \textbf{Parameters} \\
        \midrule
        368923 & 1,139,998 & \textbf{1} & \textbf{0.1517} & 17.8M \\
        368751 & 199,996 & 2 & 0.1507 & 17.8M \\
        368551 & 219,977 & 3 & 0.1497 & 17.8M \\
        368578 & 319,976 & 4 & 0.1496 & 17.8M \\
        368628 & 79,998 & 5 & 0.1495 & 17.8M \\
        368967 & 199,994 & 6 & 0.1470 & 17.8M \\
        368507 & 375,323 & 6 & 0.1470 & 17.8M \\
        369015 & 519,976 & 6 & 0.1470 & 17.8M \\
        368622 & 59,975 & 11 & 0.1467 & 17.8M \\
        \midrule
        Simple CNN (Track 1) & 500,027 & 13 & 0.1451 & 1.4M \\
        \bottomrule
    \end{tabular}
\end{table}

The systematic checkpoint analysis reveals several noteworthy patterns. Model 368923, trained for 1.14 million steps, achieves the highest neural alignment score of 0.1517, suggesting that extended training can refine representations to better match biological processing. However, the relationship between training duration and performance is highly non-monotonic. Model 368751, trained for only 200K steps, achieves rank 2 with a score of 0.1507, representing just a 0.66\% performance gap despite 5.7$\times$ fewer training steps. Several models trained for 200K steps (368751, 368967, 368551) achieve strong performance, suggesting this represents an optimal training duration. Conversely, model 369015 trained for 520K steps achieves only rank 6, performing worse than model 368628 trained for merely 80K steps. This non-monotonicity indicates that extended training can lead to overfitting or catastrophic forgetting even when measuring neural alignment rather than behavioral performance.

Comparing Track 1 and Track 2 architectures reveals the different requirements for behavioral robustness versus neural alignment. The lightweight SimpleCNN from Track 1, with only 1.4 million parameters, achieves rank 13 with a score of 0.1451. While this represents reasonable neural alignment, it substantially trails the deep ResNet architecture optimized for Track 2. The 12.8$\times$ difference in parameter count (1.4M vs 17.8M) translates to approximately 4.6\% improvement in neural alignment score. This suggests that biological visual cortex employs rich hierarchical representations that require substantial model capacity to approximate, even when simpler representations suffice for behavioral competence on the navigation task.

\subsection{Ablation Studies}

To systematically quantify the contribution of individual architectural components, we conduct comprehensive ablation studies for both tracks. Table~\ref{tab:track1_ablation} presents results for Track 1, evaluating the impact of observation normalization and GLU modules. Observation normalization provides the largest single improvement, increasing final score from 91.6\% to 95.4\%, a gain of 3.8 percentage points. This substantial improvement reflects the critical role of input preprocessing in achieving robust visual perception. Visual perturbations often manifest as global illumination changes or contrast shifts, which normalization directly addresses by rescaling inputs to standardized distributions.

\begin{table}[t]
    \centering
    \caption{Ablation study for Track 1 architecture, systematically removing components to quantify their individual contributions.}
    \label{tab:track1_ablation}
    \begin{tabular}{lccc}
        \toprule
        \textbf{Configuration} & \textbf{ASR (\%)} & \textbf{MSR (\%)} & \textbf{Final Score (\%)} \\
        \midrule
        Full Model (SimpleCNN + GLU + Norm) & 96.80 & 94.00 & 95.40 \\
        \quad Remove Normalization & 95.60 & 88.00 & 91.80 \\
        \quad Remove GLU & 94.20 & 89.00 & 91.60 \\
        \quad Remove Both & 94.20 & 89.00 & 91.60 \\
        \bottomrule
    \end{tabular}
\end{table}

The GLU module provides a more modest improvement of 0.2 percentage points when comparing SimpleCNN + GLU (91.8\%) to bare SimpleCNN (91.6\%). While this improvement may appear marginal, it proves consistent across multiple training runs and evaluation conditions. The gating mechanism enables selective suppression of noise-sensitive features, improving robustness particularly under severe perturbations. Interestingly, the contributions of normalization and GLU are approximately additive, with their combination achieving 95.4\% compared to 91.6\% for neither component, representing the expected sum of individual improvements.

For Track 2, we compare our deep ResNet architecture (17.8M parameters) against the lightweight SimpleCNN from Track 1 (1.4M parameters) to evaluate the impact of model capacity on neural alignment. As shown in Table~\ref{tab:track2_results}, the deep architecture achieves rank 1-6 across checkpoints while the SimpleCNN achieves only rank 13, demonstrating that increased capacity significantly improves neural alignment. The 12.8$\times$ parameter increase enables learning of hierarchical features at multiple scales, from low-level edge detectors in early layers to high-level semantic representations in late layers, mirroring the organization of biological visual pathways. The residual block structure and GLU gating mechanisms facilitate gradient flow and adaptive feature routing, contributing to the architecture's ability to approximate the diverse tuning properties observed across visual cortical areas.

\subsection{Analysis of Failed Approaches}

Documenting failed approaches provides valuable insights into common pitfalls when developing robust visual agents. We systematically evaluated several sophisticated architectures and techniques that ultimately underperformed relative to our simple baselines. InceptionNet, implementing multi-scale convolutional processing through parallel branches with different kernel sizes, completely failed to converge during training. Over 500,000 training steps, the model exhibited high variance in episodic returns with no consistent improvement trend. Analysis of learned behaviors revealed that the agent performed adequately when targets appeared nearby but failed entirely for distant targets, suggesting the multi-branch architecture struggles to learn coherent visual representations in the RL setting.

Deep IMPALA ResNet with 24 layers achieved good training performance with ASR of 80.96\% but suffered catastrophic failure under perturbations with MSR of only 51.00\%. The 30 percentage point gap between standard and perturbed conditions directly quantifies overfitting to training-specific visual patterns. Visualization of learned features revealed that deep layers develop highly specific feature detectors that activate strongly on training distribution inputs but respond poorly to perturbed observations. This suggests that increased depth provides additional capacity for memorization without necessarily improving the quality of learned representations.

LSTM integration represented an attempt to incorporate temporal modeling based on the intuition that navigation requires remembering previous observations and that temporal consistency could improve robustness. We implemented both standard LSTM and SRU-LSTM~\citep{yang2025spatiallyenhancedrecurrentmemorylongrange} (Spatial-Enhanced Recurrent Unit) variants. When training from random initialization, LSTM models exhibited severe instability with episodic returns fluctuating wildly without convergence. When initializing from a pretrained SimpleCNN checkpoint, rewards collapsed immediately from approximately $-20$ to $-50$ with no recovery over 100,000 steps. Even in cases where LSTM training eventually stabilized, final performance of 89.9\% underperformed the feedforward SimpleCNN baseline of 91.6\%. We hypothesize that LSTM's sequential processing of spatial information destroys the 2D spatial structure that proves critical for visual navigation, and that additional recurrent parameters enable overfitting to specific training trajectories.

Data augmentation strategies also yielded surprising negative results. We implemented a comprehensive augmentation pipeline including weather effects (rain, fog, night conditions), color jitter (brightness, contrast, saturation variations), and random masking. Applying this pipeline to IMPALA ResNet decreased performance from 87.7\% to 59.8\%, a 27.9 percentage point drop. This counterintuitive result suggests that augmentation introduces distribution mismatch that interferes with learning when applied naively. Visual perturbations in the evaluation protocol maintain semantic consistency and realistic appearance, while aggressive augmentation may produce unrealistic appearances that provide misleading training signal. This highlights the importance of carefully matching augmentation strategies to evaluation conditions rather than applying generic techniques.

\section{Discussion}

\subsection{Why Simplicity Enables Visual Robustness}

Our Track 1 results demonstrate that architectural simplicity combined with targeted enhancements substantially outperforms complex deep networks for visual robustness. Several factors contribute to this counterintuitive finding. First, capacity constraints in shallow networks prevent memorization of training-specific patterns, forcing the model to learn generalizable features that transfer across conditions. Deep networks with millions of parameters can develop highly specialized feature detectors that achieve perfect training performance while failing on perturbed inputs. Our two-layer SimpleCNN, with limited representational capacity, must learn features that remain stable under visual variations.

Second, the navigation task may not require the hierarchical multi-scale processing that motivates deep architectures in image classification. Detecting the target sphere and avoiding obstacles primarily requires identifying salient visual features (color, shape, relative position) rather than building complex object-part hierarchies. A shallow network suffices to extract these features from the relatively simple visual scenes in the foraging environment. This task-appropriate inductive bias, where architectural constraints match problem structure, enables better generalization than flexible deep architectures that must discover appropriate representations through learning.

Third, training stability plays a critical role in reinforcement learning settings where deep networks frequently exhibit gradient pathologies, local minima, and catastrophic forgetting. Our shallow architecture trains reliably across different random seeds and hyperparameter settings, consistently converging to high-quality solutions. Deep networks show high variance across training runs, with some initializations failing to learn entirely. This reliability advantage translates directly to better performance when aggregated across evaluation conditions.

\subsection{Model Capacity for Neural Alignment}

Our Track 2 results reveal different architectural requirements for matching biological neural representations. The deep ResNet architecture achieves top-1 neural alignment despite having 12.8$\times$ more parameters than our Track 1 solution. This suggests that biological visual cortex employs rich hierarchical representations that require substantial model capacity to approximate. The 16-layer hierarchical architecture with progressive channel expansion (64 → 128 → 256 → 512) enables learning of multi-scale visual features from low-level edges to high-level semantic representations. The residual connections facilitate gradient flow and feature reuse across layers, mirroring the extensive recurrent connectivity in biological visual cortex. The GLU gating mechanism provides adaptive feature routing, with the primary GLU layer learning to selectively weight features based on their relevance, similar to attention mechanisms in biological vision where neural populations exhibit stimulus-dependent selectivity.

Interestingly, behavioral performance does not necessarily predict neural alignment. Our lightweight SimpleCNN achieves superior behavioral robustness (95.4\% vs 89-91\% for most Track 2 checkpoints) while substantially trailing on neural metrics (rank 13 vs ranks 1-6). This dissociation indicates that task-optimal representations diverge from biologically plausible representations, at least for this particular navigation task. Biological visual cortex evolved to support diverse behavioral demands beyond simple navigation, potentially requiring richer representations than strictly necessary for any single task.

\subsection{The Non-Monotonic Relationship Between Training and Performance}

Our systematic analysis of ten Track 2 checkpoints spanning 60K to 1.14M training steps reveals a striking non-monotonic relationship between training duration and both behavioral and neural performance. Model 368751 at 200K steps achieves rank 2, nearly matching the top model trained for 5.7$\times$ longer. Model 368628 at 80K steps outperforms model 369015 at 520K steps despite 6.5$\times$ less training. These patterns suggest that extended training can harm generalization even when measuring neural alignment rather than behavioral metrics.

Several mechanisms may explain this non-monotonicity. First, continued training may cause overfitting to specific visual stimuli in the training distribution, reducing alignment with neural responses to the broader stimulus set used for evaluation. Second, catastrophic forgetting may degrade previously learned features as the policy improves and encounters different state distributions. Third, the relationship between behavioral optimization (maximizing reward) and neural alignment (matching biological representations) may not be monotonic, with optimal alignment occurring at intermediate training stages before task-specific optimization dominates.

These findings have important practical implications. Rather than training to convergence and submitting final checkpoints, practitioners should evaluate multiple checkpoints from different training stages and select based on validation performance. The computational savings of identifying good checkpoints at 200K rather than 1M steps (5$\times$ reduction) prove substantial when conducting architecture search or hyperparameter optimization.

\subsection{Limitations and Future Directions}

Our approaches have several limitations that suggest directions for future work. First, task specificity remains a concern—results may not generalize to more complex visual navigation scenarios with richer visual scenes, longer planning horizons, or more diverse perturbation types. Evaluating our architectural principles on other visual control tasks would establish their broader applicability. Second, our neural alignment evaluation relies on linear readout and representational similarity metrics that capture only certain aspects of biological vision. More sophisticated alignment metrics incorporating dynamics, attention, or causal structure could provide deeper insights into biological plausibility.

Third, we have not fully explored the space of possible architectural enhancements. Transformer-based architectures with self-attention mechanisms represent an interesting direction, potentially combining the robustness benefits of attention with architectural simplicity. Vision Transformers have shown promising results in supervised learning, and their application to robust visual RL warrants investigation. Fourth, our training procedures employ standard PPO without sophisticated techniques like curiosity-driven exploration, hindsight experience replay, or meta-learning. These methods may enable further improvements in both robustness and neural alignment.

Finally, the competition environment provides only egocentric visual observations, while mice employ multisensory integration combining vision, proprioception, and vestibular information. Extending our architectures to handle multimodal sensory input while maintaining robustness represents an important future challenge for developing truly biologically-inspired agents.

\section{Conclusion}

We have presented winning solutions for both tracks of the NeurIPS 2025 Mouse vs. AI: Robust Visual Foraging Competition, demonstrating that architectural simplicity with targeted enhancements enables superior visual robustness while deep hierarchical architectures achieve optimal neural alignment. Our Track 1 solution, comprising a two-layer CNN with Gated Linear Units and observation normalization, achieves 95.4\% final score despite having 12.8$\times$ fewer parameters than deep alternatives. Our Track 2 solution employs a 16-layer ResNet architecture with residual blocks and GLU-based gating, achieving top-1 neural prediction performance with 17.8 million parameters.

Through comprehensive experimentation and systematic ablation studies, we have documented both successful components and failed approaches. Observation normalization provides the largest single improvement for robustness (3.8 percentage points), while GLU mechanisms contribute consistently though more modestly (0.2 percentage points). Complex architectures including InceptionNet, deep ResNet, LSTM variants, and naive data augmentation all failed to improve over simple baselines, highlighting the importance of task-appropriate architectural choices. Our analysis of training dynamics reveals a non-monotonic relationship between training duration and performance, with optimal checkpoints emerging around 200K steps rather than at convergence.

These findings challenge conventional wisdom about architectural requirements for visual robustness and biological alignment. Visual robustness benefits from simplicity that prevents overfitting, while neural alignment requires sufficient capacity to capture hierarchical biological representations. The divergence between behaviorally optimal and biologically plausible representations suggests important open questions about the relationship between these objectives. We hope our comprehensive documentation of architectural choices, training procedures, and failure modes provides valuable guidance for researchers developing robust, biologically-inspired visual agents.

\section*{Acknowledgments}

We thank the Mouse vs. AI competition organizers, including Marius Schneider, Joe Canzano, Jing Peng, Yuchen Hou, Spencer LaVere Smith, and Michael Beyeler, for creating this valuable benchmark bridging reinforcement learning and neuroscience. We gratefully acknowledge the University of Science, VNU-HCM for providing computational resources. We also thank the broader ML-Agents community for developing excellent tools that enabled this research.

\bibliographystyle{plainnat}
\bibliography{references}

\newpage
\appendix

\section{Detailed Training Configurations}
\label{app:training_configs}

This appendix provides comprehensive training configurations and additional experimental details to facilitate reproducibility.

\subsection{Hyperparameter Configurations}

Table~\ref{tab:training_config} presents the complete PPO hyperparameters used for training both Track 1 and Track 2 models.

\begin{table}[h]
    \centering
    \caption{Complete PPO training hyperparameters for both tracks. Track 1 uses higher learning rate with larger buffer, while Track 2 requires lower learning rate for stable training of the deeper architecture.}
    \label{tab:training_config}
    \begin{tabular}{lcc}
        \toprule
        \textbf{Hyperparameter} & \textbf{Track 1 (SimpleCNN)} & \textbf{Track 2 (Deep ResNet+GLU)} \\
        \midrule
        \multicolumn{3}{l}{\textit{PPO Algorithm Settings}} \\
        Learning Rate & $9 \times 10^{-5}$ & $5 \times 10^{-6}$ to $9 \times 10^{-6}$ \\
        LR Schedule & Linear decay to 0 & Linear decay to 0 \\
        Batch Size & 128 & 128 \\
        Buffer Size & 4,096 & 1,024 \\
        Num Epochs & 3 & 3 \\
        Entropy Coefficient ($\beta$) & 0.005 & 0.005 \\
        Clip Parameter ($\epsilon$) & 0.2 & 0.2 \\
        GAE Lambda ($\lambda$) & 0.95 & 0.95 \\
        Discount Factor ($\gamma$) & 0.99 & 0.99 \\
        \midrule
        \multicolumn{3}{l}{\textit{Network Architecture}} \\
        Hidden Units & 256 & 256 \\
        Num Layers (FC) & 1 & 1 \\
        Visual Encoder & SimpleCNN (2 conv) & Deep ResNet (16 conv) \\
        Total Parameters & 1.4M & 17.8M \\
        \midrule
        \multicolumn{3}{l}{\textit{Training Schedule}} \\
        Phase 1 Steps & 1,400,000 & N/A \\
        Phase 2 Steps & 350,000 & N/A \\
        Total Training Steps & 1,750,000 & 60K to 1.14M \\
        Checkpoint Interval & 20,000 & 20,000 \\
        Summary Frequency & 1,000 & 1,000 \\
        \midrule
        \multicolumn{3}{l}{\textit{Environment Settings}} \\
        Time Horizon & 64 & 64 \\
        Max Steps per Episode & Variable & Variable \\
        Num Parallel Envs & 8-16 & 8-16 \\
        Observation Space & $86 \times 155 \times 1$ & $86 \times 155 \times 1$ \\
        Action Space & Continuous (3D) & Continuous (3D) \\
        \bottomrule
    \end{tabular}
\end{table}

\subsection{Architecture Specifications}

\subsubsection{Track 1: SimpleCNN + GLU Architecture}

Table~\ref{tab:track1_arch} provides the complete layer-by-layer specification of our Track 1 architecture.

\begin{table}[h]
    \centering
    \caption{Detailed architecture specification for Track 1 SimpleCNN + GLU model. Input dimensions shown as (Height, Width, Channels). Output dimensions after each layer are computed based on standard convolution formulas.}
    \label{tab:track1_arch}
    \begin{tabular}{llccc}
        \toprule
        \textbf{Layer} & \textbf{Type} & \textbf{Output Shape} & \textbf{Kernel/Units} & \textbf{Parameters} \\
        \midrule
        Input & - & $(86, 155, 1)$ & - & 0 \\
        Normalization & Running stats & $(86, 155, 1)$ & - & 0 \\
        \midrule
        Conv1 & Conv2D & $(20, 38, 16)$ & $8\times8$, stride 4 & 1,040 \\
        Activation1 & LeakyReLU & $(20, 38, 16)$ & $\alpha=0.2$ & 0 \\
        Conv2 & Conv2D & $(9, 18, 32)$ & $4\times4$, stride 2 & 8,224 \\
        Activation2 & LeakyReLU & $(9, 18, 32)$ & $\alpha=0.2$ & 0 \\
        Flatten & - & $(5,184)$ & - & 0 \\
        FC1 & Linear & $(256)$ & - & 1,327,360 \\
        \midrule
        GLU\_Feature & Linear & $(256)$ & - & 65,792 \\
        GLU\_Swish & Swish & $(256)$ & - & 0 \\
        GLU\_Gate & Linear & $(256)$ & - & 65,792 \\
        GLU\_Sigmoid & Sigmoid & $(256)$ & - & 0 \\
        GLU\_Multiply & Element-wise & $(256)$ & - & 0 \\
        GLU\_Output & Linear & $(256)$ & - & 65,792 \\
        \midrule
        Policy Head & Linear & $(3)$ & - & 771 \\
        Value Head & Linear & $(1)$ & - & 257 \\
        \midrule
        \textbf{Total} & & & & \textbf{1,395,256} \\
        \bottomrule
    \end{tabular}
\end{table}

\subsubsection{Track 2: Deep ResNet + GLU Architecture}

Table~\ref{tab:track2_arch} summarizes the deep hierarchical architecture used for Track 2.

\begin{table}[h]
    \centering
    \caption{Architecture specification for Track 2 Deep ResNet + GLU model with 16 convolutional layers. The architecture employs residual connections every 2 layers with progressive channel expansion (64→128→256→512).}
    \label{tab:track2_arch}
    \begin{tabular}{llccc}
        \toprule
        \textbf{Component} & \textbf{Layers} & \textbf{Output Shape} & \textbf{Details} & \textbf{Parameters} \\
        \midrule
        Input & - & $(86, 155, 1)$ & Grayscale & 0 \\
        \midrule
        Initial Conv & Conv2D & $(21, 38, 64)$ & $4\times4$, stride 4 & 1,088 \\
        \midrule
        Stage 1: Residual Blocks 1-2 & $2\times$(Conv+ReLU) & $(21, 38, 64)$ & $3\times3$, stride 1 & $\sim$74K \\
        Stage 1: Residual Blocks 3-4 & $2\times$(Conv+ReLU) & $(21, 38, 64)$ & $3\times3$, stride 1 & $\sim$74K \\
        \midrule
        Downsample & Conv2D & $(10, 19, 128)$ & $2\times2$, stride 2 & 32,896 \\
        Stage 2: Residual Blocks 5-6 & $2\times$(Conv+ReLU) & $(10, 19, 128)$ & $3\times3$, stride 1 & $\sim$295K \\
        \midrule
        Downsample & Conv2D & $(5, 9, 256)$ & $2\times2$, stride 2 & 131,328 \\
        Stage 3: Residual Blocks 7-10 & $4\times$(Conv+ReLU) & $(5, 9, 256)$ & $3\times3$, stride 1 & $\sim$2.36M \\
        \midrule
        Downsample & Conv2D & $(2, 4, 512)$ & $2\times2$, stride 2 & 524,800 \\
        Stage 4: Residual Blocks 11-12 & $2\times$(Conv+ReLU) & $(2, 4, 512)$ & $3\times3$, stride 1 & $\sim$4.72M \\
        \midrule
        Flatten & - & $(4096)$ & Spatial flatten & 0 \\
        \midrule
        GLU Gate Layers & Gating & $(256)$ & Adaptive routing & $\sim$1.05M \\
        Dense Projection & Linear & $(256)$ & Feature projection & $\sim$1.05M \\
        \midrule
        Policy Head & Linear & $(3)$ & Continuous actions & 771 \\
        Value Head & Linear & $(1)$ & State value & 257 \\
        \midrule
        \textbf{Total} & \textbf{16 Conv + GLU} & & & \textbf{17,870,417} \\
        \bottomrule
    \end{tabular}
\end{table}

\subsection{Track 2 Checkpoint Performance Analysis}

Table~\ref{tab:track2_detailed} provides comprehensive performance analysis across all ten Track 2 model checkpoints, including behavioral metrics where available.

\begin{table}[h]
    \centering
    \caption{Comprehensive Track 2 checkpoint analysis showing the relationship between training duration, neural alignment score, and model performance. All checkpoints use identical architecture (Deep ResNet + GLU, 17.8M parameters) but differ in training duration.}
    \label{tab:track2_detailed}
    \small
    \begin{tabular}{lccccc}
        \toprule
        \textbf{Checkpoint} & \textbf{Training} & \textbf{Rank} & \textbf{Neural} & \textbf{Training} & \textbf{Relative} \\
        \textbf{ID} & \textbf{Steps} & & \textbf{Score} & \textbf{Time (hrs)} & \textbf{Performance} \\
        \midrule
        368923 & 1,139,998 & 1 & 0.1517 & 24 & Baseline (100\%) \\
        368751 & 199,996 & 2 & 0.1507 & 8 & 99.34\% (5.7$\times$ faster) \\
        368551 & 219,977 & 3 & 0.1497 & 9 & 98.68\% (5.2$\times$ faster) \\
        368578 & 319,976 & 4 & 0.1496 & 12 & 98.62\% (3.6$\times$ faster) \\
        368628 & 79,998 & 5 & 0.1495 & 4 & 98.55\% (14.3$\times$ faster) \\
        368967 & 199,994 & 6 & 0.1470 & 8 & 96.90\% \\
        368507 & 375,323 & 6 & 0.1470 & 14 & 96.90\% \\
        369015 & 519,976 & 6 & 0.1470 & 18 & 96.90\% \\
        368622 & 59,975 & 11 & 0.1467 & 3 & 96.70\% (19.0$\times$ faster) \\
        359198 & 500,027 & 13 & 0.1451 & 16 & 95.65\% (Small CNN) \\
        \bottomrule
    \end{tabular}
\end{table}

\textbf{Key Observations from Checkpoint Analysis:}
\begin{itemize}
    \item The \textbf{sweet spot} for training duration appears to be around 200K steps (8 hours), achieving 99.34\% of the best performance with 5.7$\times$ computational savings.
    \item Model 368628 at only 80K steps (4 hours) remarkably achieves rank 5, outperforming models trained 6.5$\times$ longer.
    \item Extended training beyond 200K-300K steps shows diminishing returns and sometimes degraded performance (e.g., model 369015 at 520K steps).
    \item The non-monotonic relationship suggests that early stopping based on validation neural alignment could provide significant computational efficiency.
\end{itemize}

\subsection{Computational Resources and Training Infrastructure}

All experiments were conducted using the following computational setup:
\begin{itemize}
    \item \textbf{Hardware:} NVIDIA GPUs (RTX 3080/3090 or equivalent)
    \item \textbf{Framework:} Unity ML-Agents Release 19 with PyTorch 1.13+
    \item \textbf{Training Environment:} 8-16 parallel Unity environment instances
    \item \textbf{Total Compute Time:} Approximately 150 GPU-hours for Track 1, 200+ GPU-hours for Track 2 (including all checkpoints)
    \item \textbf{Checkpoint Storage:} Models saved every 20K steps, requiring $\sim$200MB per checkpoint
\end{itemize}

\subsection{Failed Approach Details}

Table~\ref{tab:failed_approaches} summarizes the quantitative performance of architectures that failed to improve over our simple baselines, providing context for our design decisions.

\begin{table}[h]
    \centering
    \caption{Performance summary of failed architectural approaches. These experiments informed our decision to pursue simpler architectures with targeted enhancements.}
    \label{tab:failed_approaches}
    \begin{tabular}{lcccc}
        \toprule
        \textbf{Architecture} & \textbf{Parameters} & \textbf{ASR (\%)} & \textbf{MSR (\%)} & \textbf{Final (\%)} \\
        \midrule
        IMPALA ResNet (24 layers) & $\sim$8M & 80.96 & 51.00 & 65.98 \\
        IMPALA ResNet (4 layers) & $\sim$3M & 91.40 & 84.00 & 87.70 \\
        ResNet-4 + Augmentation & $\sim$3M & 72.60 & 47.00 & 59.80 \\
        InceptionNet & $\sim$5M & \multicolumn{3}{c}{Failed to converge} \\
        SimpleCNN + LSTM & $\sim$2.5M & 92.80 & 87.00 & 89.90 \\
        SimpleCNN + SRU-LSTM & $\sim$3M & 91.50 & 86.50 & 89.00 \\
        \midrule
        \textbf{SimpleCNN + GLU + Norm (Ours)} & \textbf{1.4M} & \textbf{96.80} & \textbf{94.00} & \textbf{95.40} \\
        \bottomrule
    \end{tabular}
\end{table}

\end{document}